\documentclass[runningheads,a4paper]{llncs}

\usepackage{amsmath,graphicx}
\usepackage{epstopdf}
\usepackage{amssymb}
\usepackage [english]{babel}
\usepackage [autostyle, english = american]{csquotes}
\usepackage{color,soul}
\MakeOuterQuote{"}
\makeatletter
\@ifundefined{showcaptionsetup}{}{%
	\PassOptionsToPackage{caption=false}{subfig}}
\usepackage{subfig}
\makeatother
\newcommand{\minus}{\scalebox{0.75}[1.0]{$-$}}

\begin{document}
\mainmatter
\title{Multiple Reflection Symmetry Detection via Linear-Directional Kernel Density Estimation}
\titlerunning{Multiple Reflection Symmetry Detection}
\author{Mohamed Elawady\inst{1}\and Olivier Alata\inst{1} \and Christophe Ducottet\inst{1} \and C\'{e}cile Barat\inst{1} \and Philippe Colantoni\inst{2}}
\institute{Universit\'{e} Jean Monnet, CNRS, UMR 5516, Laboratoire Hubert Curien, F-42000, Saint-Etienne, France,\\
	\email{mohamed.elawady@univ-st-etienne.fr},
	\and
	Universit\'{e} Jean Monnet, Centre Interdisciplinaire d'Etudes et de Recherches sur l'Expression Contemporaine n\textsuperscript{0} 3068, Saint-\'{E}tienne, France
}
\authorrunning{Mohamed Elawady et al.}
\maketitle
\begin{abstract}
Symmetry is an important composition feature by investigating similar sides inside an image plane. It has a crucial effect to recognize man-made or nature objects within the universe. Recent symmetry detection approaches used a smoothing kernel over different voting maps in the polar coordinate system to detect symmetry peaks, which split the regions of symmetry axis candidates in inefficient way. We propose a reliable voting representation based on weighted linear-directional kernel density estimation, to detect multiple symmetries over challenging real-world and synthetic images. Experimental evaluation on two public datasets demonstrates the superior performance of the proposed algorithm to detect global symmetry axes respect to the major image shapes.
\end{abstract}
\section{Introduction}

Reflection symmetry is a fundamental principle of visual perception to feel the equally distributed weights within foreground objects inside an image. These weights are inspected respect to textural complexity of their shapes, in such non-identical manner to preserve a well-balanced composition between similar objects and their surrounding background \cite{hobbs1995visual,Freeman2007,Zakia2010}. Detection of reflection symmetry has a principal intermediate-level role in recent computer vision applications \cite{Zhao2014,ram2016vehicle,yuan2016incremental,abdolali2016automatic}. Liu et al. \cite{Liu2010} described the global symmetry as top-tier visual features, which are distributed uniformly across the image sides and contributed to define an uppermost similarity behavior. This paper focuses on detecting multiple bilateral symmetry axes by exploring the geometrical correlation between regions on a global scale.

The baseline algorithm was proposed in 2006 by Loy and Eklundh \cite{Loy2006}. They analyzed the bilateral symmetry from image features' constellation by introducing the general scheme: (1) detection of local feature points (i.e. SIFT), associated with local geometrical properties (location, orientation, scale) and descriptor vectors. (2) pairwise matching and evaluation of a local symmetry magnitude of their descriptors, to generate symmetry candidates. (3) accumulation of their symmetry magnitude in a Hough-like voting space parametrized with orientation and displacement, to identify the dominant reflection axes inside an image. The first survey of symmetry detection algorithms was introduced by the computer vision group of Pennsylvania State University in 2008 \cite{Park2008}. The same group conducted symmetry detection challenges in 2011 \cite{Rauschert2011} and 2013 \cite{Liu2013}, where the baseline algorithm \cite{Loy2006} still outperformed the participated approaches \cite{Mo2011,Kondra2013,Michaelsen2013,Patraucean2013}. Other keypoint-based algorithms \cite{Cho2009,Cai2014} also proposed feature refinement techniques for better results. Edge/contour-based features \cite{Ming2013,Cicconet2014,Wang2015,Atadjanov2015,Cicconet2016,Atadjanov2016,Elawady2016} are modernly used instead of the intensity-based, due to saliency properties in detecting well-defined symmetric structures inside an object.
In both approaches, a limited number of feature points are detected in the image, axis candidates are randomly sampled across the voting space. These sparse symmetry candidates further need to be grouped through a smoothing kernel to define relevant mono- or multi- axis hypothesis. Our idea is to formulate the voting problem as a density estimation problem, by computing the probability of detecting symmetry axis at every position and orientation inside the image plane. 

Kernel density estimation is one of the most popular techniques in nonparametric statistics. Density estimates are controlled by a smoothing bandwidth and a weighting kernel function. Density estimates with linear kernels have been introduced in 1954 \cite{akaike1954approximation}, and then have been adapted to deal with directional data since the mid 1980s \cite{Hall1987}. Many computer vision applications used kernel density estimation for linear data \cite{elgammal2002background,mittal2004motion,zhang2005applying,cremers2006kernel,van2008kernel,wang2009semi,tavakoli2011fast,liu2012unsupervised}, and fewer recently used it for directional data \cite{vuollo2016analyzing,pardo2017directional}. Garcia-Portugues et al. \cite{Garcia-Portugues2013} derived the general principle of joint kernel density estimator for linear-directional data. 

Our contribution is twofold. First we propose a weighted joint density estimator to handle both orientation and displacement information. Second, we introduce a robust linear-directional kernel-based voting representation for reflection symmetry detection. This approach is evaluated for multiple symmetry detection using two public datasets. The remaining sections of this paper are organized as follows. Section 2 describes the proposed algorithm. Sections 3 and 4 present the experimental details and results on two public datasets. Finally, the conclusion is given in section 5.
\section{Algorithm Details}

\begin{figure}[ht!]
    \centering
    \includegraphics[width=1.0\textwidth]{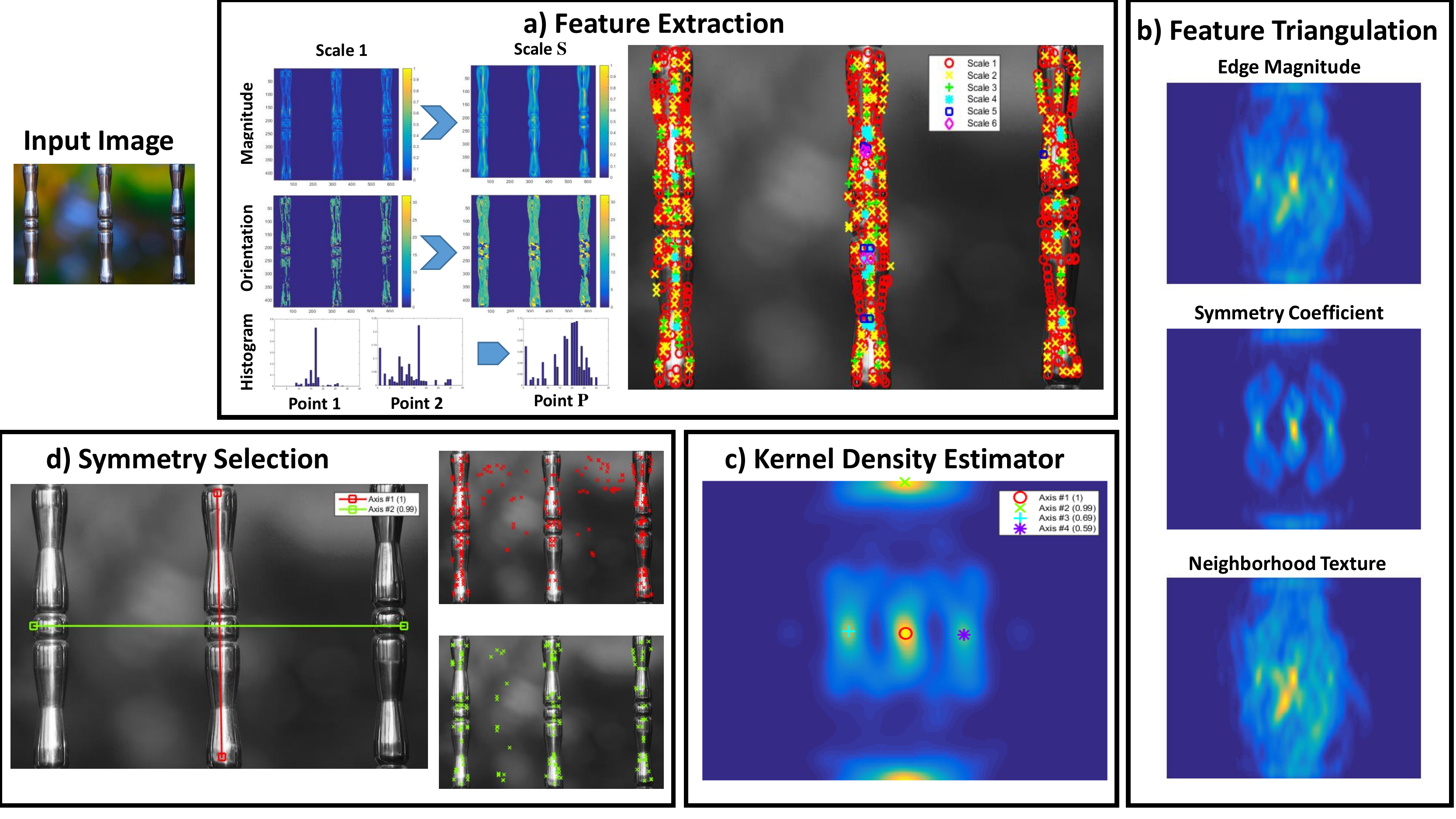}
    \caption{The proposed framework of reflection symmetry detection, using a weighted linear-directional kernel density estimator (LD-KDE). Best seen on screen (zoom-in for details).}
    \label{fig:framework}
\end{figure}

Given an image, our algorithm focuses globally to detect all symmetry axes using a dense and regular estimation of linear-directional density, as briefly shown in figure \ref{fig:framework}. First, we extract wavelet-based features with different scales, accompanied with edge and textural characteristics (for better display, only features with high magnitude are displayed over the gray-scale version of the input image). Second, we triangulate each feature pair at each scale with respect to the origin of the feature space, in order to define symmetrical weights in the polar coordinate system. Third, we formulate a voting representation based on weighted pairs via linear-directional kernel density estimation. Finally, the global symmetry axes are well-chosen by searching for maximum peaks, and spatially defined by the convex hull of the voting features.

\subsection{Feature Extraction and Normalization}

Upon the application of Morlet wavelet over an image with multiple scales, feature points  $\{ p^i = [p^i_x,p^i_y]^T,\: i=1 \ldots P\}$ are sampled, as detailed in \cite{Elawady2016}, along a regular grid with respect to image size (width $W$ and height $H$), along side with their local edge components (scale $\sigma^i$, maximum wavelet response $J^i$ and direction $\tau^i$ over all orientations) plus neighboring textural histograms $h^i$. These points are normalized with keeping aspect ratio as following:
\begin{equation}
\hat{p}^{i} = \frac{p^{i}-c_{W,H}}{max(W,H)}
\end{equation}
where $c_{W,H}$ represents the original image center ($\frac{W}{2}$, $\frac{H}{2}$). So that, the feature space is transferred from the dynamic-sized image plane $[1,W]\times[1,H]$ to a unified version $[\minus 1,1]\times[\minus 1,1]$.

\subsection{Pairwise Symmetry Triangulation}

\begin{figure}[ht!]
    \centering
    \includegraphics[width=0.8\textwidth]{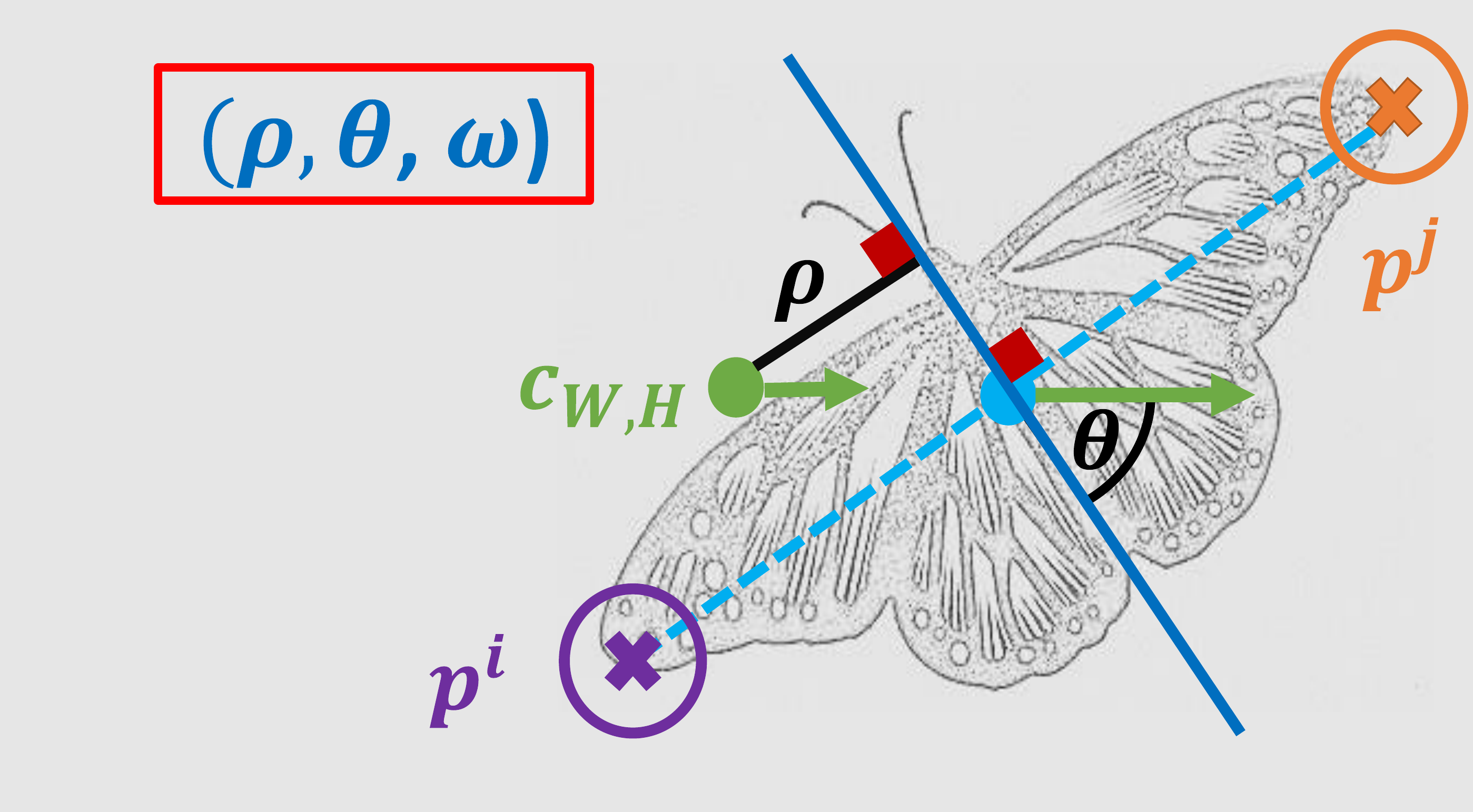}
    \caption{Symmetric triangulation for a feature pair. Best seen on screen.}
    \label{fig:SymTri}
\end{figure}

We first define a set of feature pairs $\{q_n=(\hat{p}^{i},\hat{p}^{j})\; |\; n=1,\ldots,N\}$ such that  $i \neq j$ and $\sigma^i=\sigma^j$. Then, we perform a triangulation process (as illustrated in figure~\ref{fig:SymTri}) with respect to the feature origin, producing the symmetry candidate axis as the bisector of the pair segment. The  candiate axis is parametrized by angle $\theta_n \in [0,\pi]$, and displacement $\rho_n \in [-\frac{\sqrt[]{2}}{2},\frac{\sqrt[]{2}}{2}]$) and has a symmetry weight $\omega_n$ \cite{Cicconet2014,Elawady2016}  defined as follows:
\begin{equation}
	\omega_n = m_n c_n d_n, \; ||\omega||_1 = 1
\label{eq:weights}
\end{equation}
where $m_n \in [0,1]$ is a semi-dense edge magnitude, $c_n \in [0,1]$ is a mirror symmetry coefficient based on the local edge orientation of points of the pair, $d_n \in [0,1]$ is a measure based on histograms $h^i$ and $h^j$, representing  the similarity between the local texture around the feature points, and $||.||_1$ is $L_1$ norm.

\subsection{Weighted Linear-Directional Kernel Density Estimation}
One dimensional linear variable $\rho$ represents displacement part of a candidate axis, assuming $\rho_1, \ldots, \rho_N$ samples of $\rho$ with size $N$. Let $\mu$ describes two dimensional directional variable (circular data corresponds to angle $\theta$, representing orientation part of a candidate axis, see figure~\ref{fig:SymTri}), assuming $\mu_1,\ldots,\mu_N$ samples of $\mu$ with the same size of $\rho$.
Inspired from \cite{Parzen1962}, the linear kernel density estimator $f_l(.)$ is defined as
\begin{equation}
f_l(x;g) = \frac{1}{Ng} \sum_{n=1}^{N}G(\frac{x-\rho_n}{g}), \: x \in \mathbb{R}
\end{equation}
\begin{equation}
G(u) = \frac{1}{(2\pi)^{\frac{1}{2}}} e^{-\frac{1}{2}|u|^2},
\end{equation}
where $G(.)$ is a Gaussian kernel with bandwidth parameter $g$. Inspired from \cite{Hall1987}, the directional kernel density estimator $f_d(.)$ is defined as:
\begin{equation}
f_d(y;k) = C(k) \sum_{n=1}^{N} L(y^T\mu_n;k), \: y \in \Omega_2
\end{equation}
\begin{equation}
L(x;k) = e^{kx}, \: C(k) = \frac{1}{2 \pi S(0,k)},
\end{equation}
\begin{equation}
y=[cos(\theta),sin(\theta)], \: \mu_n=[cos(\theta_n),sin(\theta_n)],
\end{equation}
where $L(.)$ is a von-Mises Fisher kernel \cite{Mardia2009} with concentration parameter $k$, and normalization constant $C(k)$. $S(.)$ is the modified Bessel function of the first kind. $y$ is remarked as directional unit-vector of angle $\theta$, such that $||y||=1$.

As axis candidate samples $(\rho_1,\mu_1),\ldots,(\rho_N,\mu_N)$ are associated with symmetry weights $\omega_1,\omega_2,\cdots,\omega_N$, and use of the linear-directional density estimator $f_{l,d}(.)$ in \cite{Garcia-Portugues2013}. We define the extended weighted version $\hat{f}_{l,d}(.)$ as:
\begin{equation}
\hat{f}_{l,d}(x,y;g,k) = \frac{C(k)}{Ng}  \sum_{n=1}^{N} \omega_n G(\frac{x-\rho_n}{g}) L(y^T\mu_n;k) 
\end{equation}
assuming that linear and directional data are independent resulting dot product between accompanying kernels. 
Previous weights (see equation~\ref{eq:weights}) are multiplied by $N$ in order to normalize $\hat{f}_{l,d}$.
\newline

\begin{figure}[tbph]
	\centering
	\subfloat[]{\includegraphics[width=0.5\textwidth]{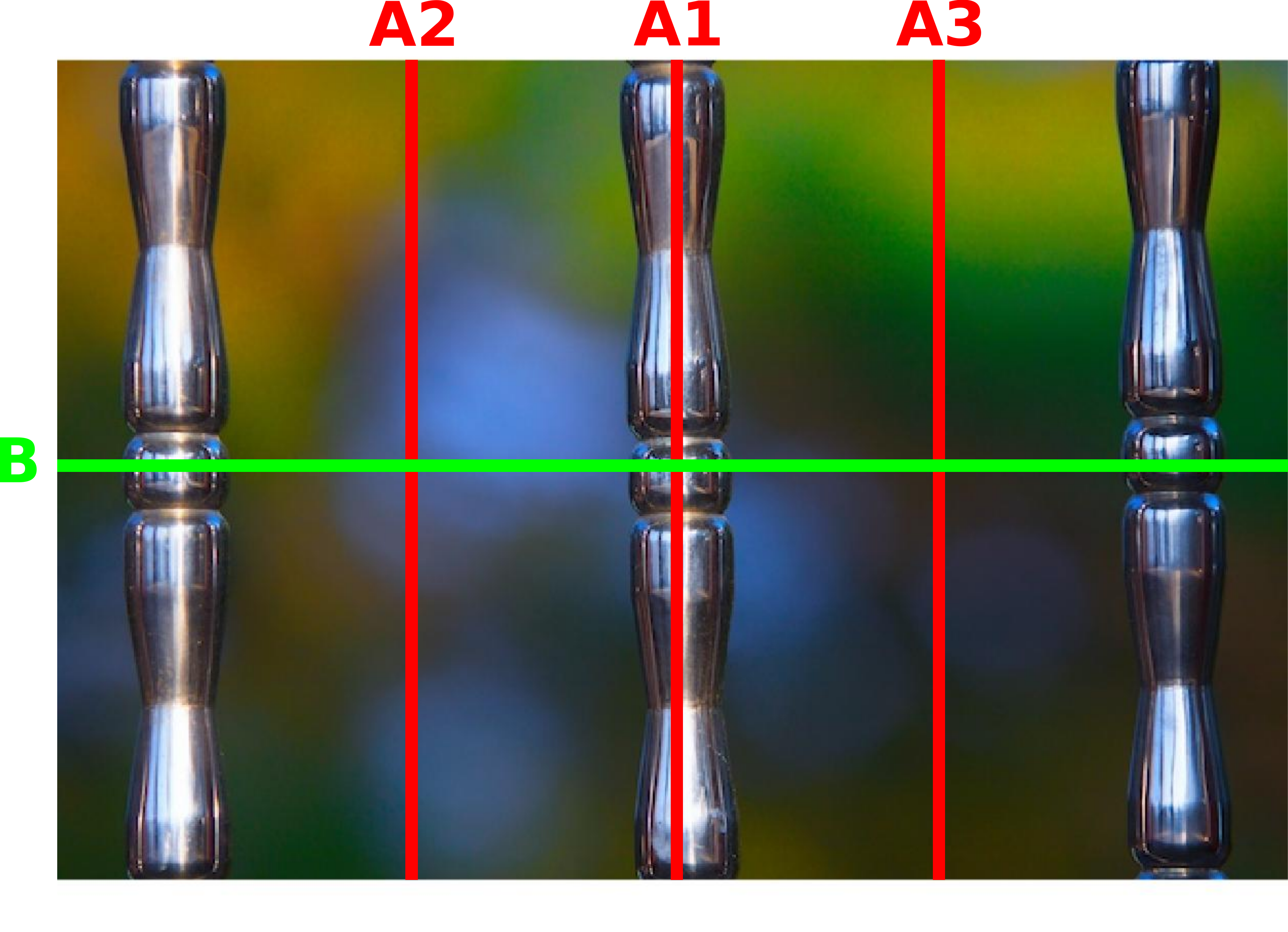}\label{fig:KDE_Input} } 
	\subfloat[]{\includegraphics[width=0.5\textwidth]{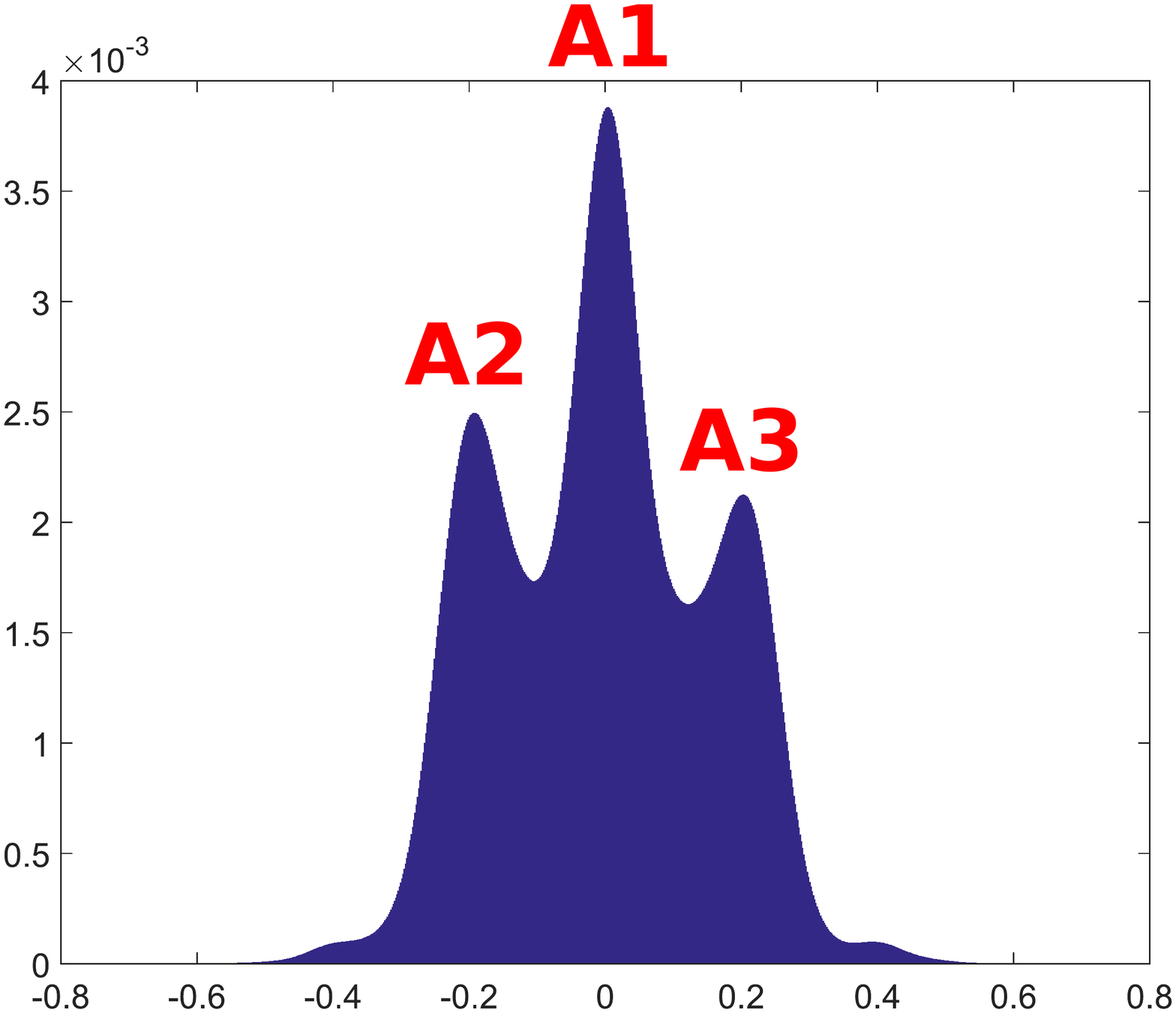}\label{fig:KDE_1Dd} }
	\\
\subfloat[]{\includegraphics[width=0.5\textwidth]{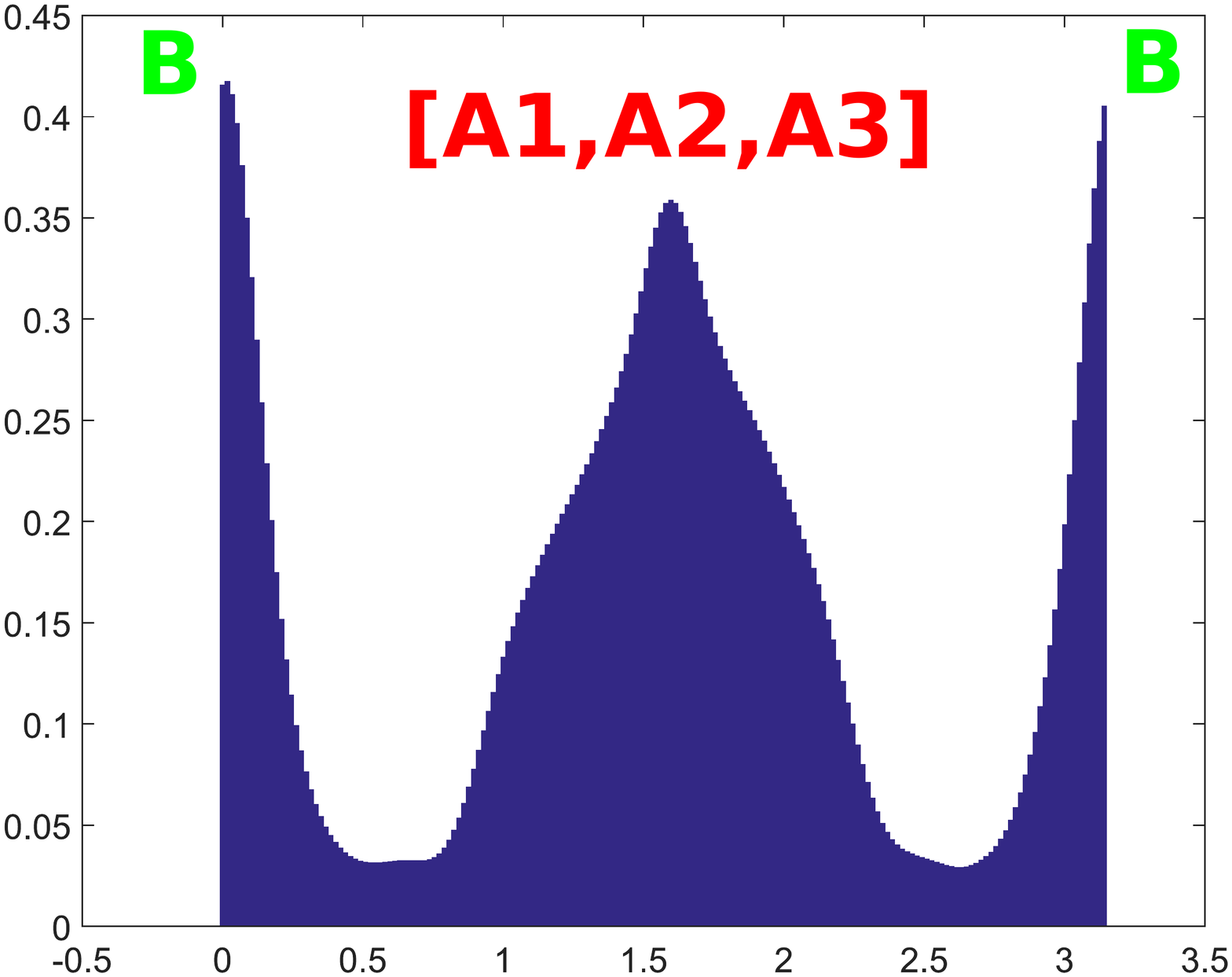}\label{fig:KDE_1Do} } 
	\subfloat[]{\includegraphics[width=0.5\textwidth]{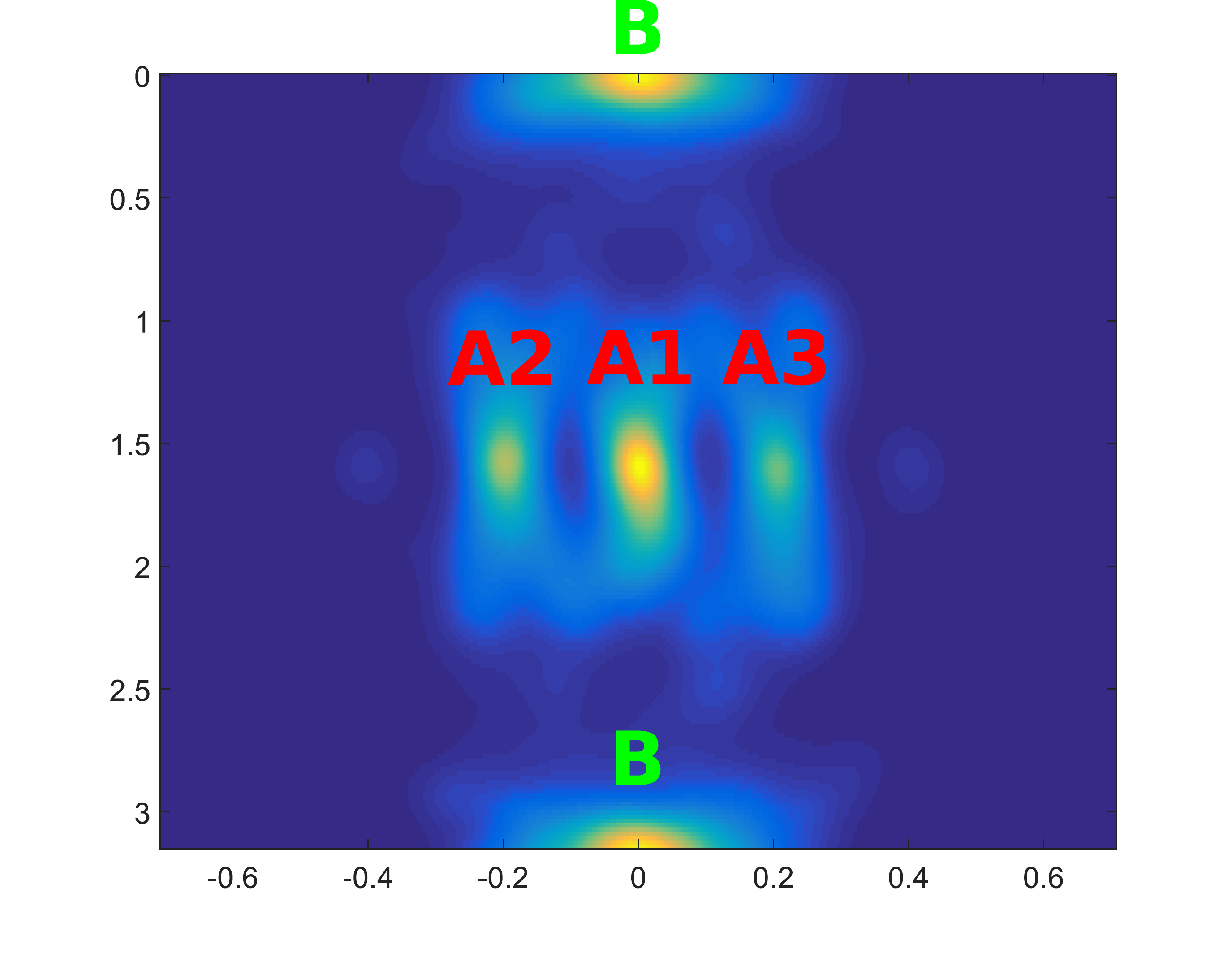}\label{fig:KDE_2D} }
	\caption{Symmetry detection process: (a) Input image with global symmetry axis candidates. (b) The output of weighted linear kernel density $\hat{f}_{l}(x;g)$ over $800$ bins. (c) The output of the weighted directional kernel density $\hat{f}_{d}(y;k)$ over $180$ bins. (d) The output of the weighted linear-directional kernel density $\hat{f}_{l,d}(x,y;g,k)$ over $800 \times 180$ bins. Maximal peaks are associated with global symmetry axes. Best seen on screen.}
	\label{fig:KDE}
\end{figure}

The multiple symmetry peaks inside the voting representation $\hat{f}_{l,d}(.)$ are identified through finding non-interleaved extreme spots via a standard non-maximal suppression technique \cite{canny1986computational}. The spatial extent of each peak representing a symmetry axis is determined by the convex hull of the voting pair associated with the peak \cite{Loy2006}. Figures [\ref{fig:KDE_Input}-\ref{fig:KDE_2D}] present an example of multiple symmetry detection, using 1D and 2D kernel-based voting maps. Three vertical symmetry axes are shown in the weighted linear kernel density $\hat{f}_{l}(x;g)$ (figure \ref{fig:KDE_1Dd}). Two major directional axes appear in the weighted directional kernel density $\hat{f}_{d}(y;k)$ at angles $\theta = 90^{\circ}, 180^{\circ}$ (figure \ref{fig:KDE_1Do}). All global symmetry axes are clearly recognized through the combination version of the previous weighted densities $\hat{f}_{l,d}(x,y;g,k)$ (figure \ref{fig:KDE_2D}). To obtain such representation, as $\theta$ originally belongs to $[0,\pi)$, each angle value is multiplied by $2$ in order to obtain an appropriate periodicity with the directional kernel.  

\section{Implementation and Evaluation Details}
We compare our reflection symmetry detection approach against three different methods (Loy2006 \cite{Loy2006}, Cicconet2014 \cite{Cicconet2014}, and Elawady2016 \cite{Elawady2016}). We executed their source codes with default parameter values, assigned by the authors for stable performance. For our approach, we empirically set the linear-kernel bandwidth parameter $g$ to $0.03$, and directional-kernel concentration parameter $k$ to $40$. Two public datasets are used to represent multiple reflection symmetry detection results: (1) PSU dataset: Liu's vision group proposed a symmetry groundtruth for Flickr images (\# images = 142, \# symmetries = 479) in ECCV2010\footnote[1]{http://vision.cse.psu.edu/research/symmetryCompetition/index.shtml}, CVPR2011\footnote[2]{http://vision.cse.psu.edu/research/symmComp/index.shtml} and CVPR2013\footnote[3]{http://vision.cse.psu.edu/research/symComp13/content.html}. Non-duplicative images are combined from three previously mentioned versions for challenging comparisons. (2) NY dataset: Cicconet et al. \cite{Cicconet2016} presented a new symmetry database (\# images = 63, \# symmetries = 188) in 2016\footnote[4]{http://symmetry.cs.nyu.edu/}, providing more accurate and consistent groundtruth for multiple symmetry endpoints.  

Quantitative comparisons are performed as proposed in \cite{Rauschert2011,Liu2013}, where a detected symmetry axis considered as a true positive (TP): 1) The angle between the detected symmetry axis and its corresponding groundtruth symmetry axis is less than 10 degree; 2) The distance between the centers of detected and same groundtruth axes is less than 20\% minimum length of the axes. Multiple detections can match to the same ground-truth axis, but not vice versa. The overall performance of algorithms are defined through the precision and recall rates:
\begin{equation}
precision = \frac{TP}{TP+FP}, \: recall = \frac{TP}{TP+FN}.
\end{equation}
where false positive (FP) are non-matched detected axes with any groundtruth, false negative (FN) are non-matched groundtruth with any detected axes. Close detections are clustered as one proposal axis to avoid duplicates in true positive or false positive calculations.
\section{Experiments and Discussions}
\begin{figure}[tbph]
	\centering
	\subfloat[PSUm]{\includegraphics[width=0.5\textwidth]{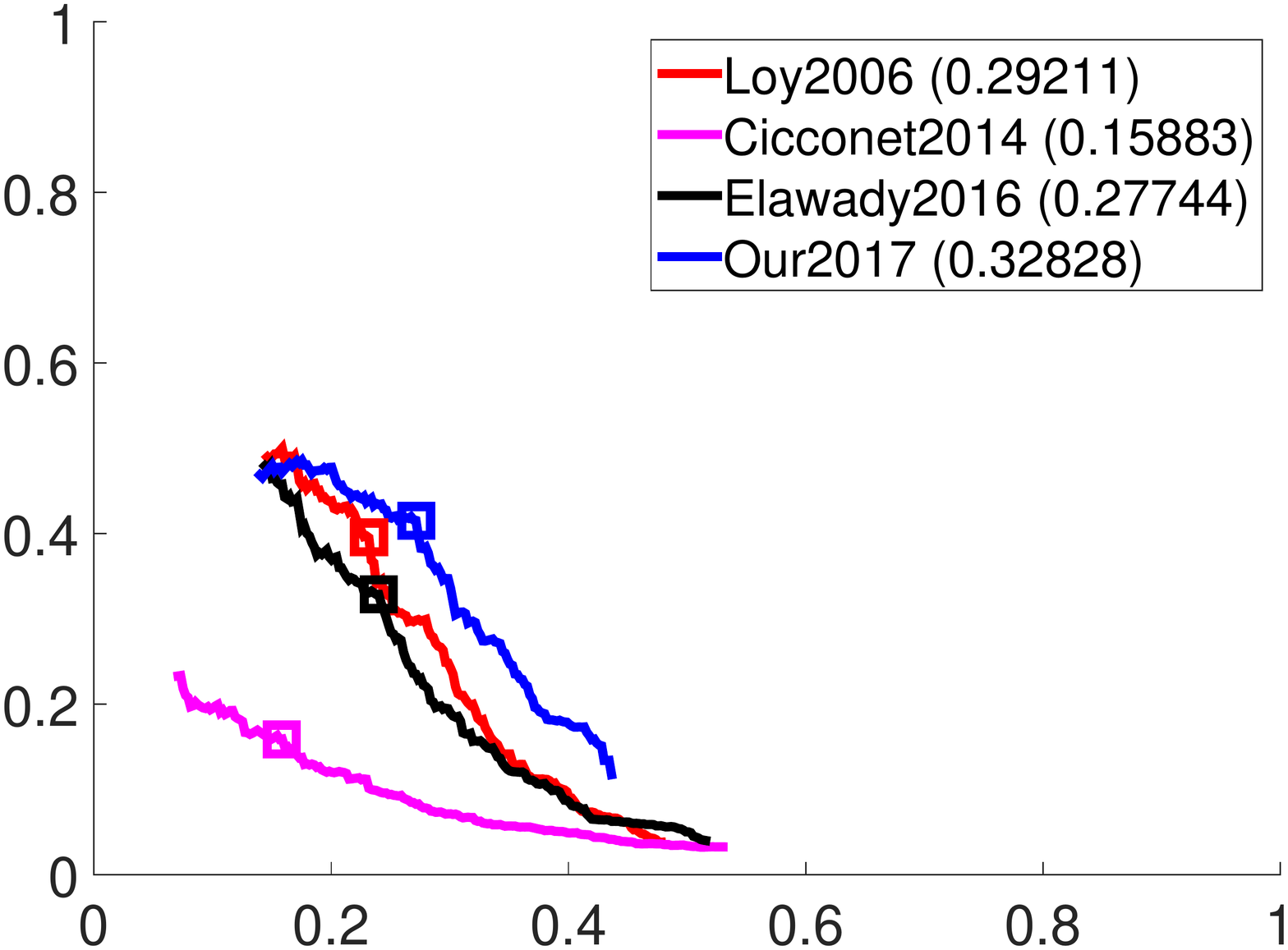}\label{fig:prCurve_PSUm} } 
	\subfloat[NYm]{\includegraphics[width=0.5\textwidth]{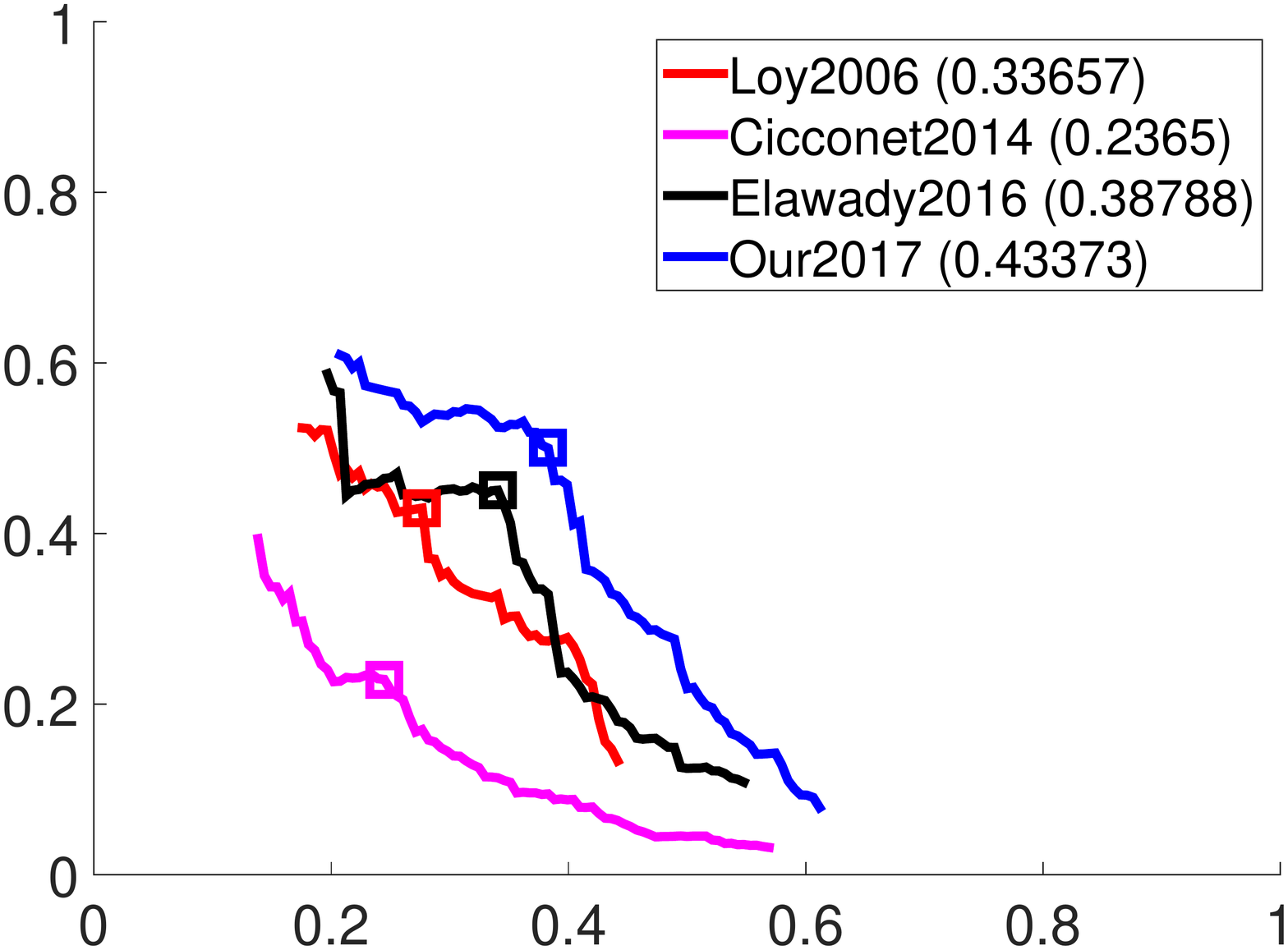}\label{fig:prCurve_NYm} }
	\caption{Precision-Recall curves on (a)PSUm \cite{Rauschert2011,Liu2013} and (b)NYm \cite{Cicconet2016} datasets to show the superior performance of our method "Our2017" against the three prior algorithms ("Loy2006" \cite{Loy2006}, "Cic2014" \cite{Cicconet2014}, "Ela2016" \cite{Elawady2016}). The maximum F1 scores are qualitatively presented as square symbols along the curves, and quantitatively indicated between parentheses inside the top-right legends. Best seen on screen.}
	\label{fig:prCurve}
\end{figure}

\begin{figure}[tbph]
	\centering
	\includegraphics[width=1\textwidth]{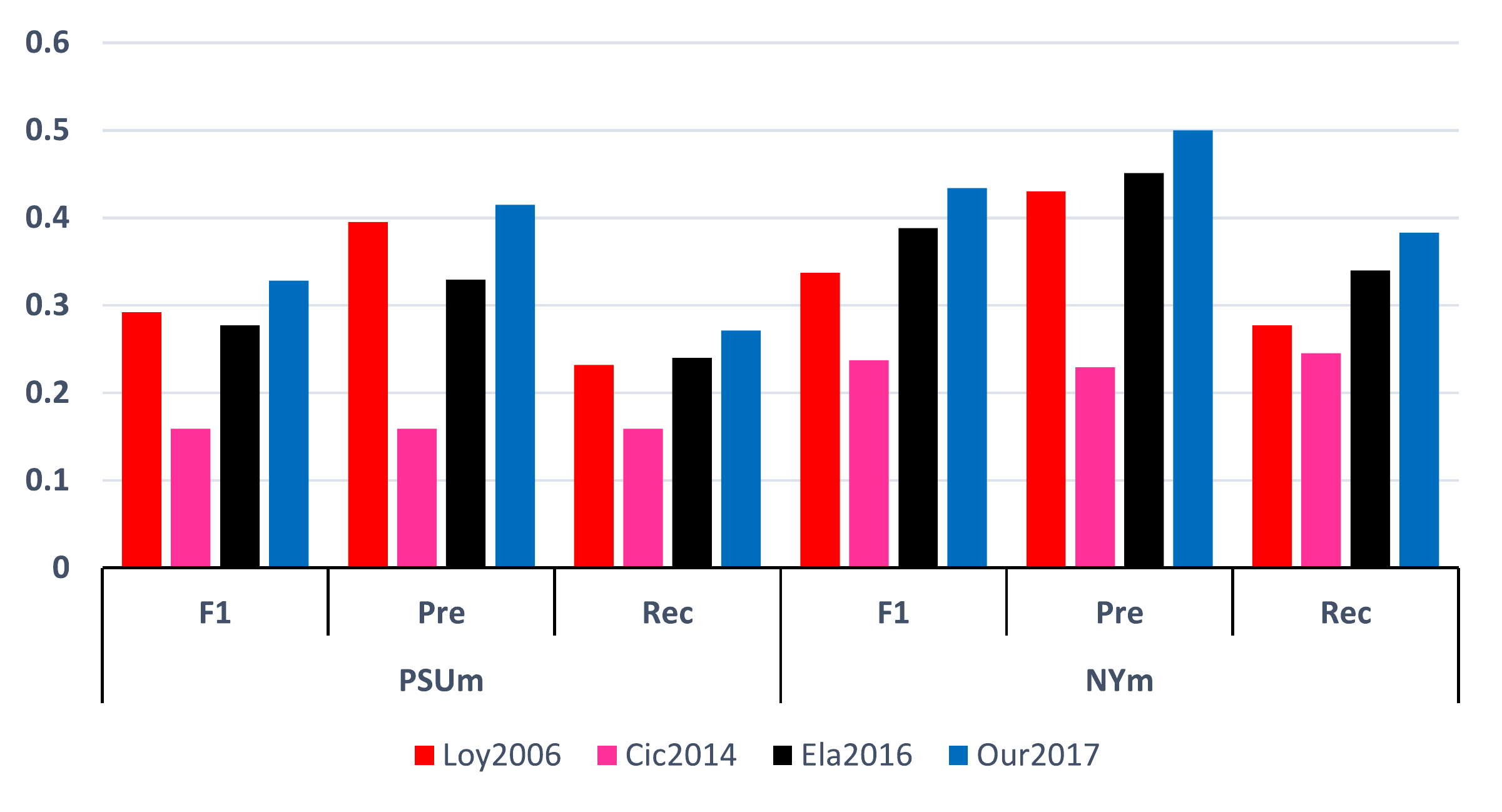}
	\caption{Comparison of maximum F1 score and its equivalent precision and recall rates among two different dataset (PSUm \cite{Rauschert2011,Liu2013}, NYm \cite{Cicconet2016}) for our method "Our2017" against the baseline algorithm "Loy2006" \cite{Loy2006} and two of the recent algorithms "Cic2014" \cite{Cicconet2014} and "Ela2016" \cite{Elawady2016}.}
	\label{fig:MulSymStat}
\end{figure}

\begin{figure}
	\centering
    \subfloat[ref\_rm\_79 - GT]{\includegraphics[width=0.33\textwidth]{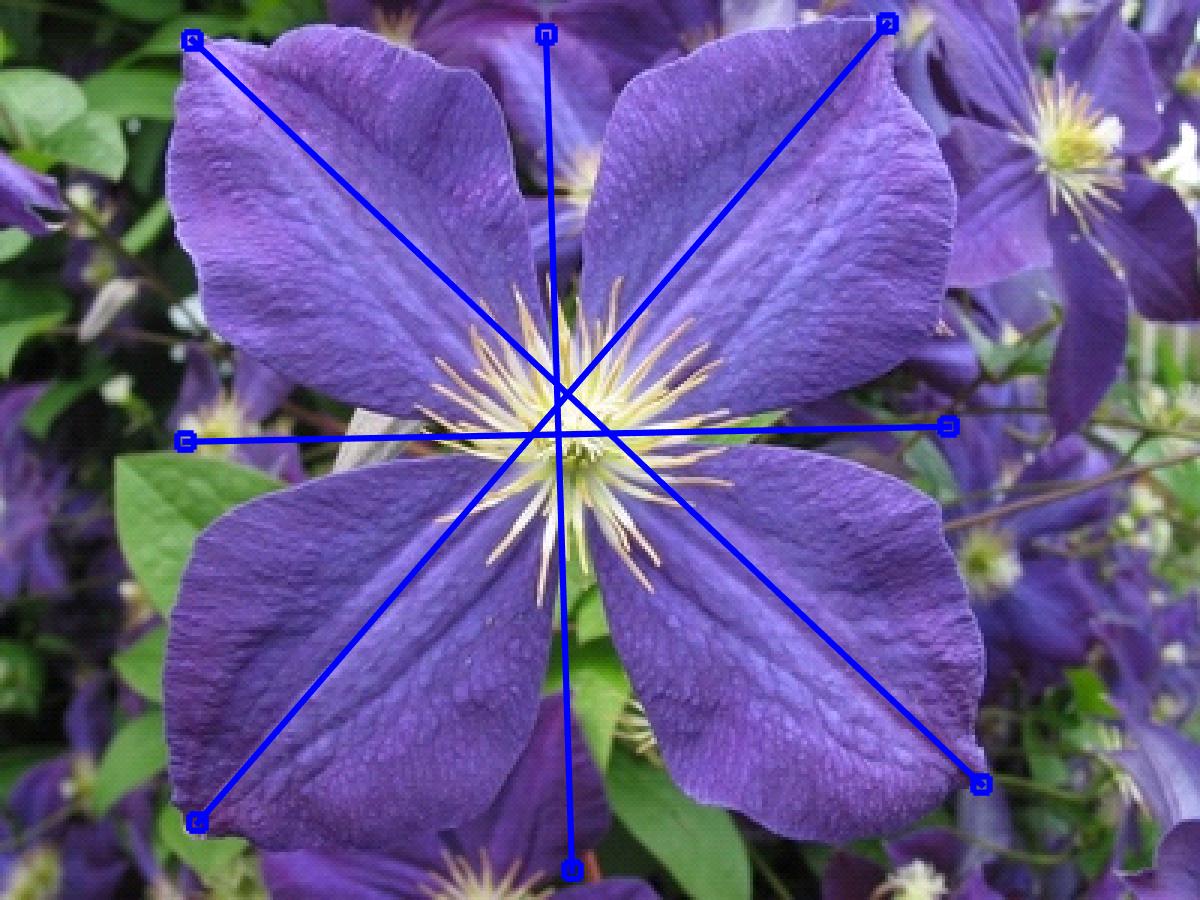}\label{fig:PSU1_GT} } 
    \subfloat[ref\_rm\_71 - GT]{\includegraphics[width=0.33\textwidth]{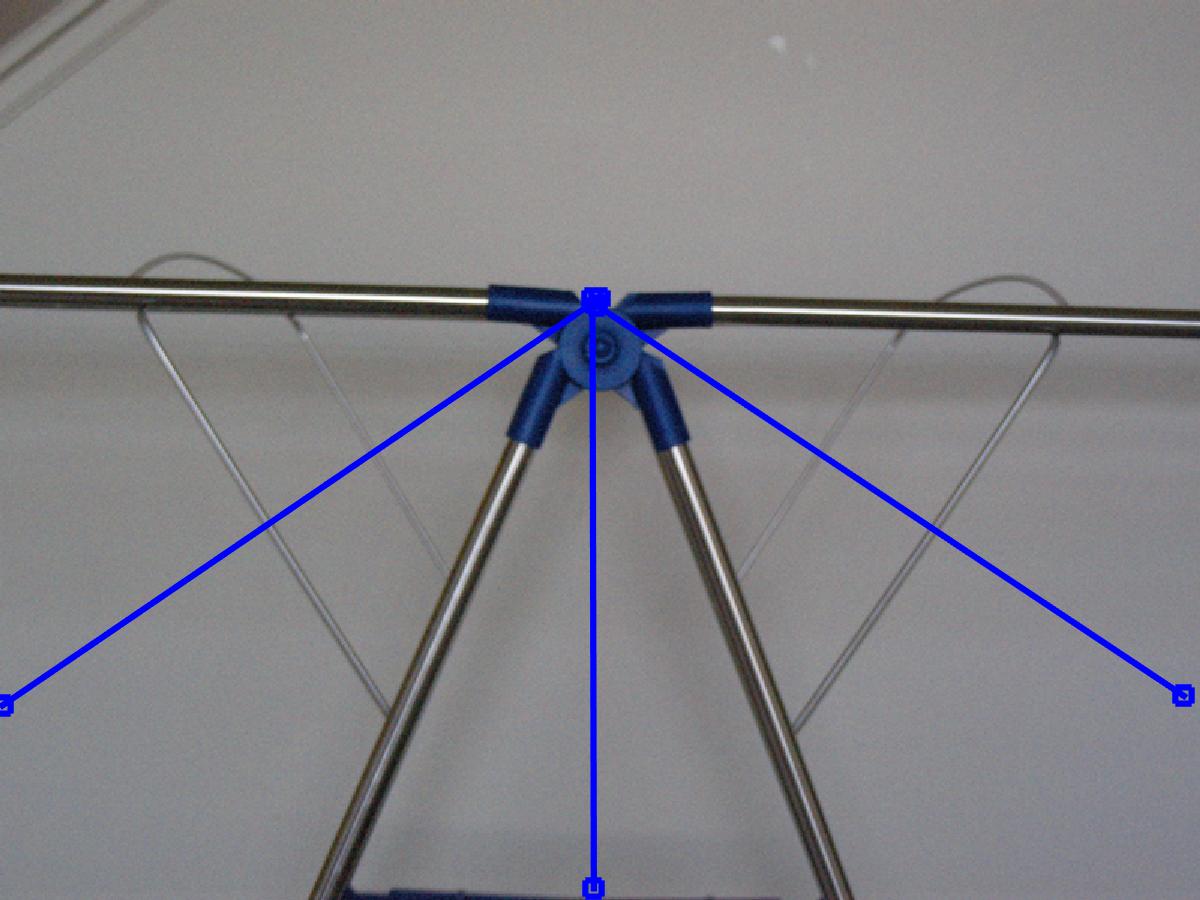}\label{fig:PSU2_GT} } 
	\subfloat[I012 - GT]{\includegraphics[width=0.33\textwidth]{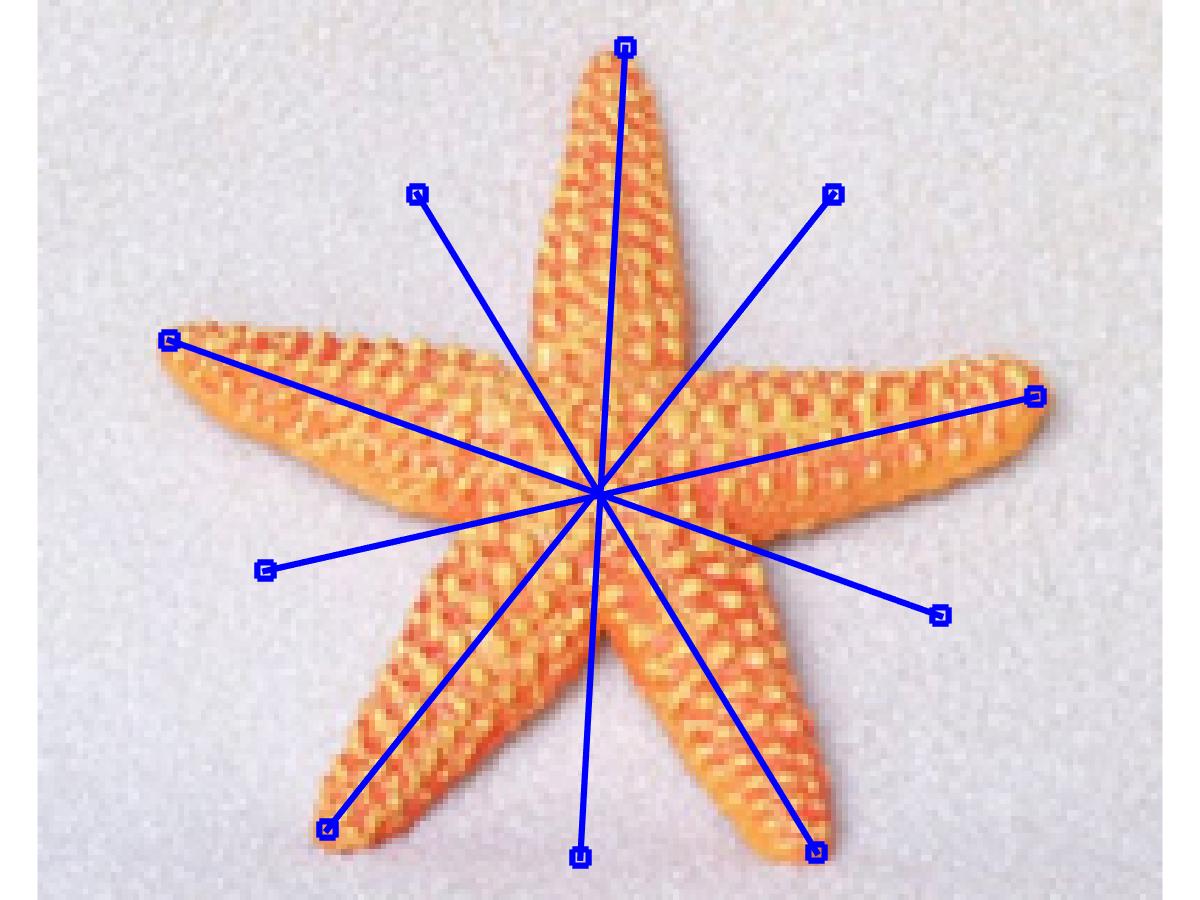}\label{fig:NY1_GT} }
    	\\
	\subfloat[ref\_rm\_79 - Our2017]{\includegraphics[width=0.33\textwidth]{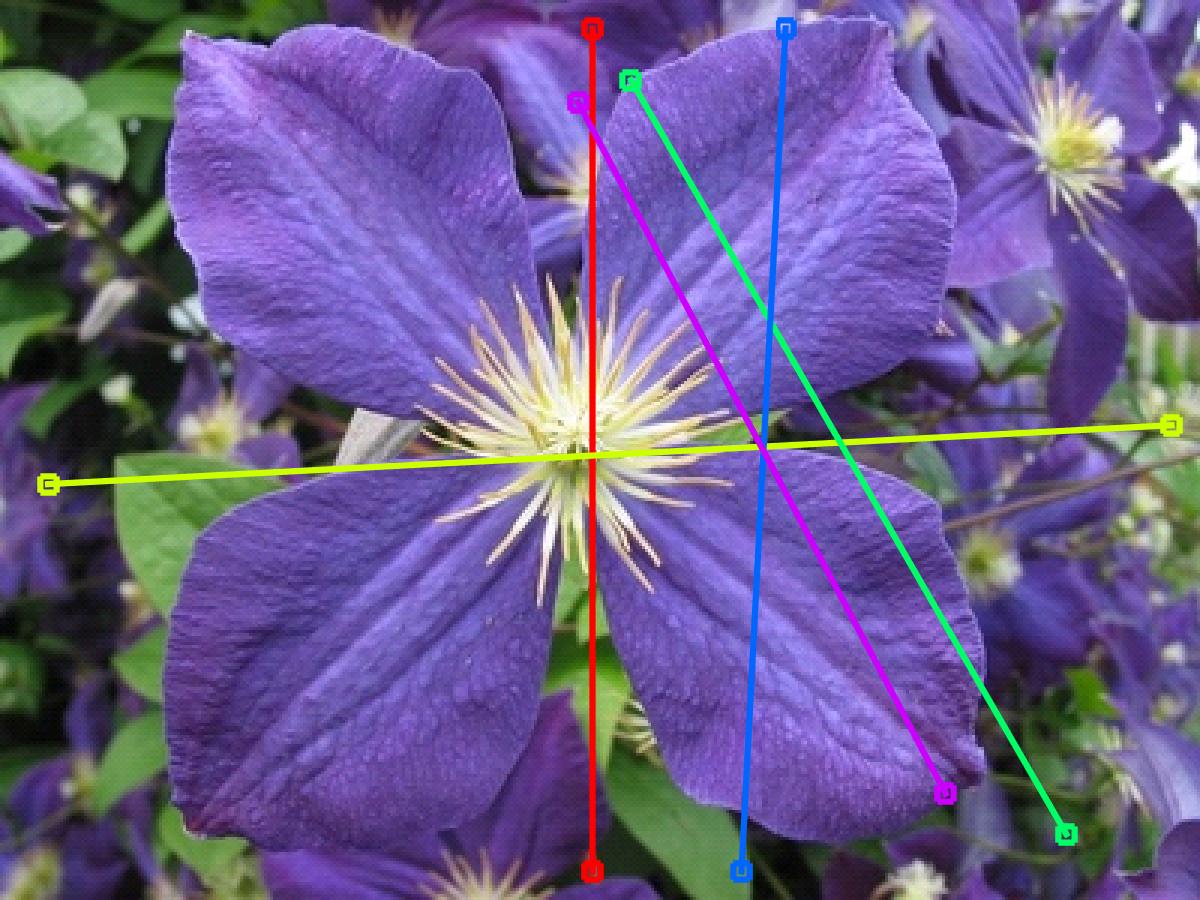}\label{fig:PSU1_Our} } 
    \subfloat[ref\_rm\_71 - Our2017]{\includegraphics[width=0.33\textwidth]{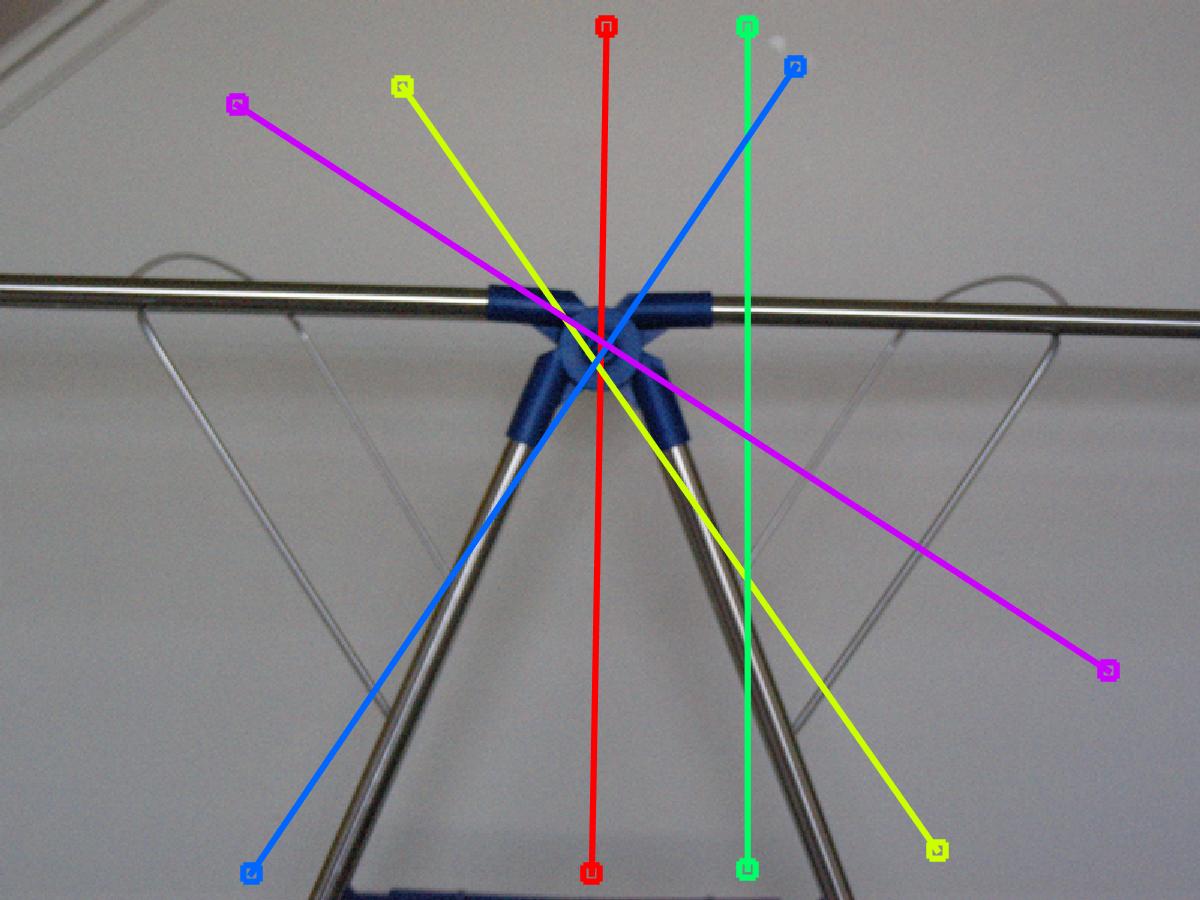}\label{fig:PSU2_Our} }
	\subfloat[I012 - Our2017]{\includegraphics[width=0.33\textwidth]{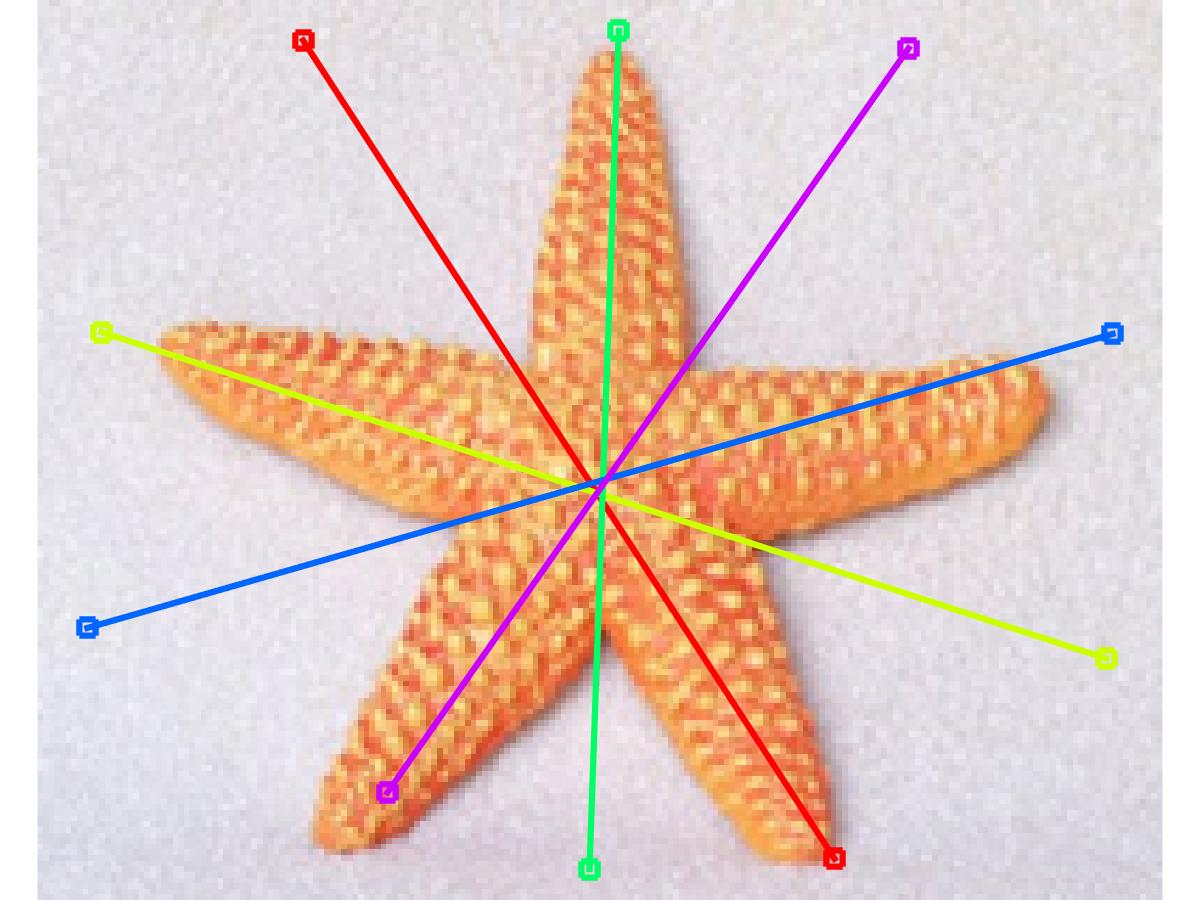}\label{fig:NY1_Our} } 
    \\	
    \subfloat[ref\_rm\_79 - Ela2016]{\includegraphics[width=0.33\textwidth]{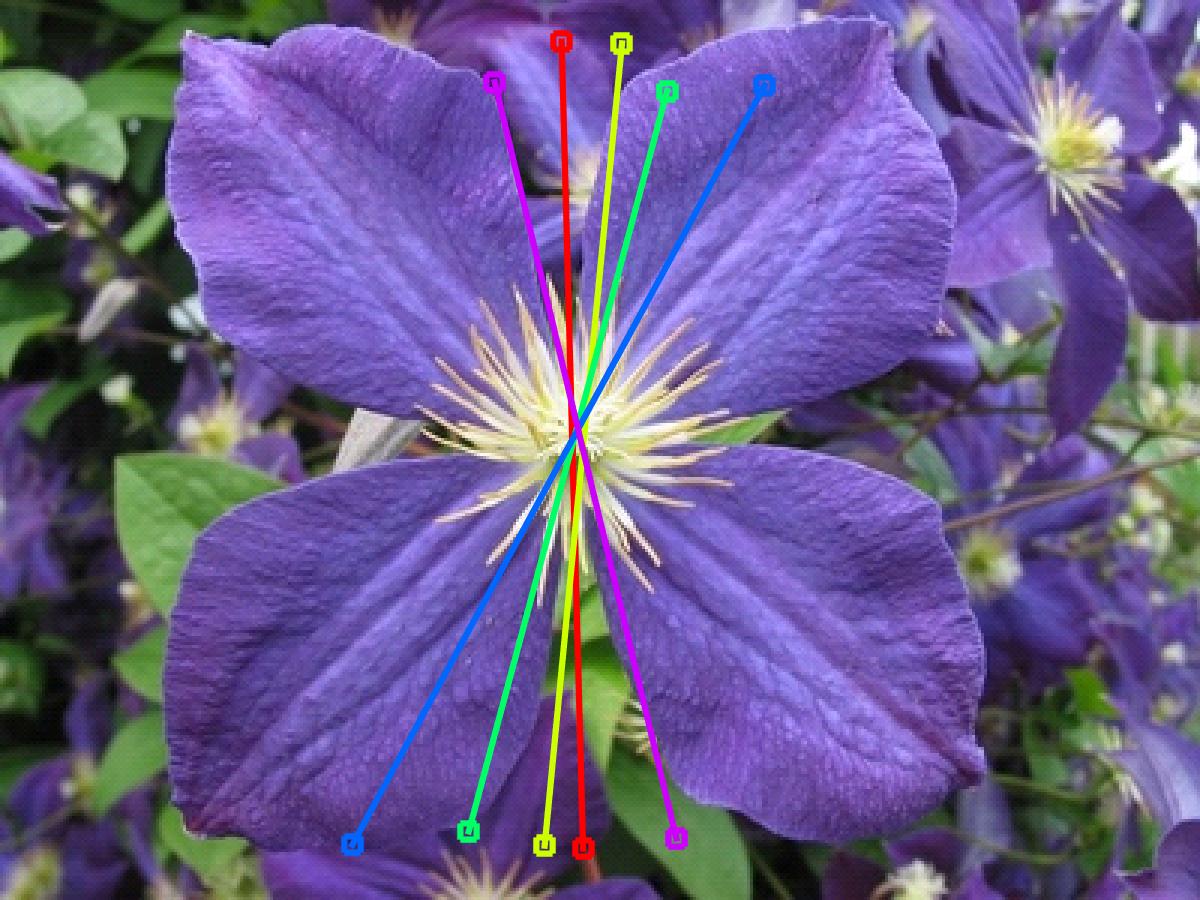}\label{fig:PSU1_Ela} }
    \subfloat[ref\_rm\_71 - Ela2016]{\includegraphics[width=0.33\textwidth]{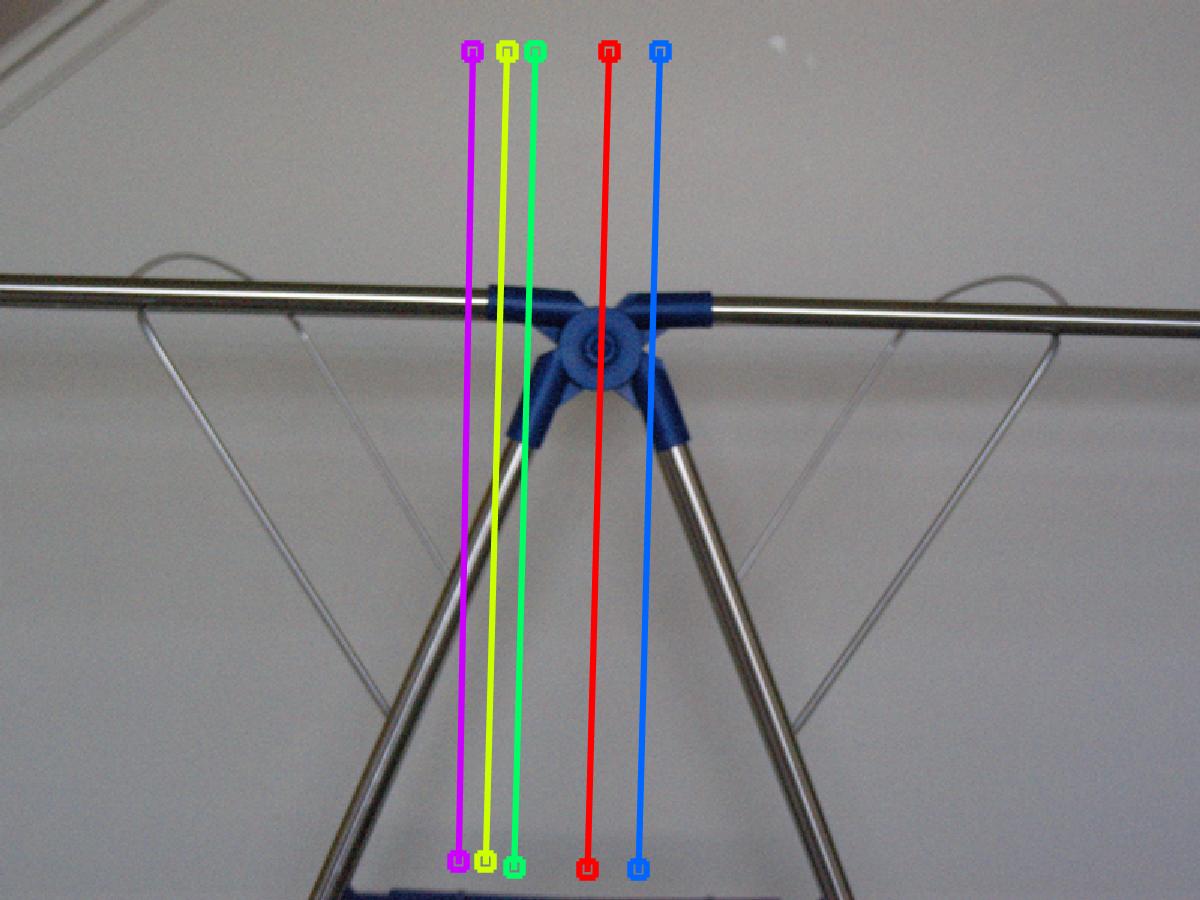}\label{fig:PSU2_Ela} }
	\subfloat[I012 - Ela2016]{\includegraphics[width=0.33\textwidth]{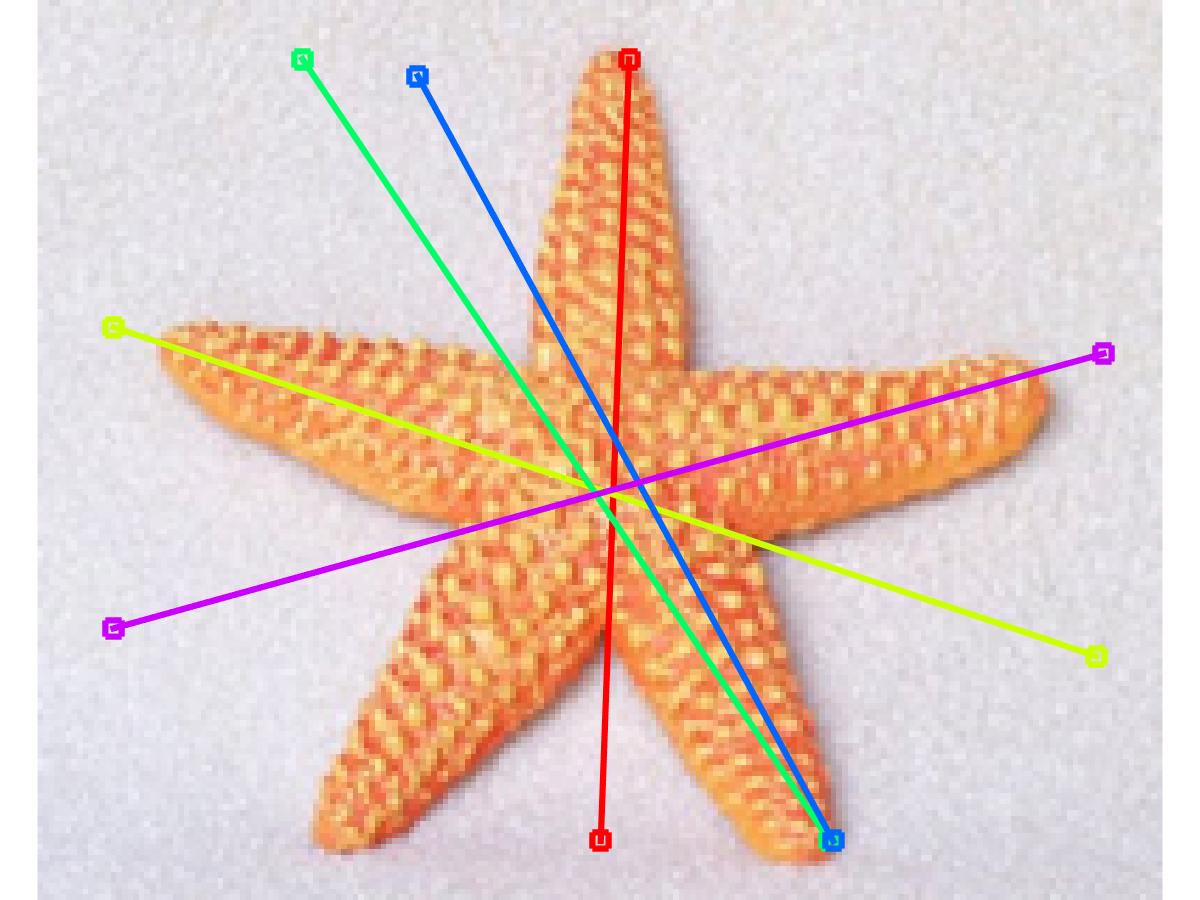}\label{fig:NY1_Ela} } 
	\\
	\subfloat[ref\_rm\_79 - Loy2006]{\includegraphics[width=0.33\textwidth]{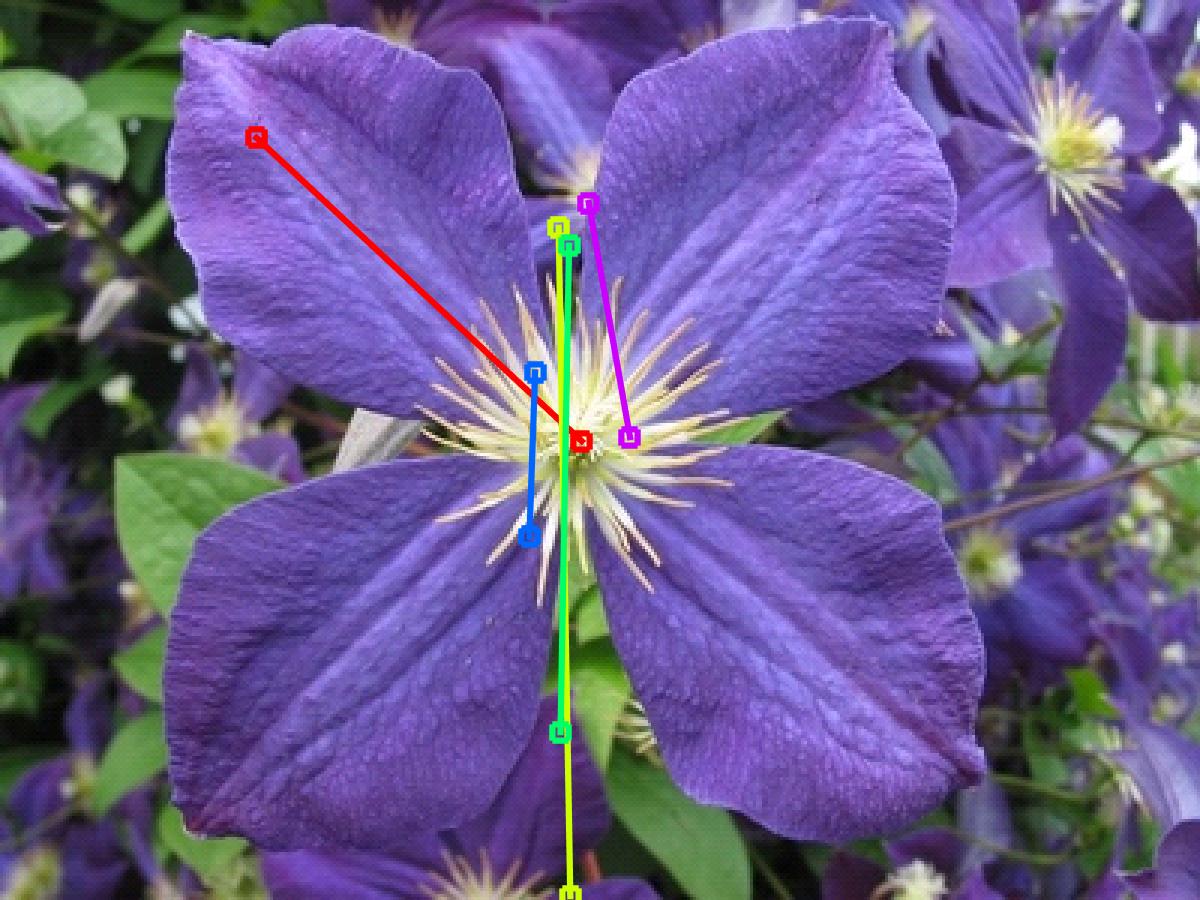}\label{fig:PSU1_Loy} }
\subfloat[ref\_rm\_71 - Loy2006]{\includegraphics[width=0.33\textwidth]{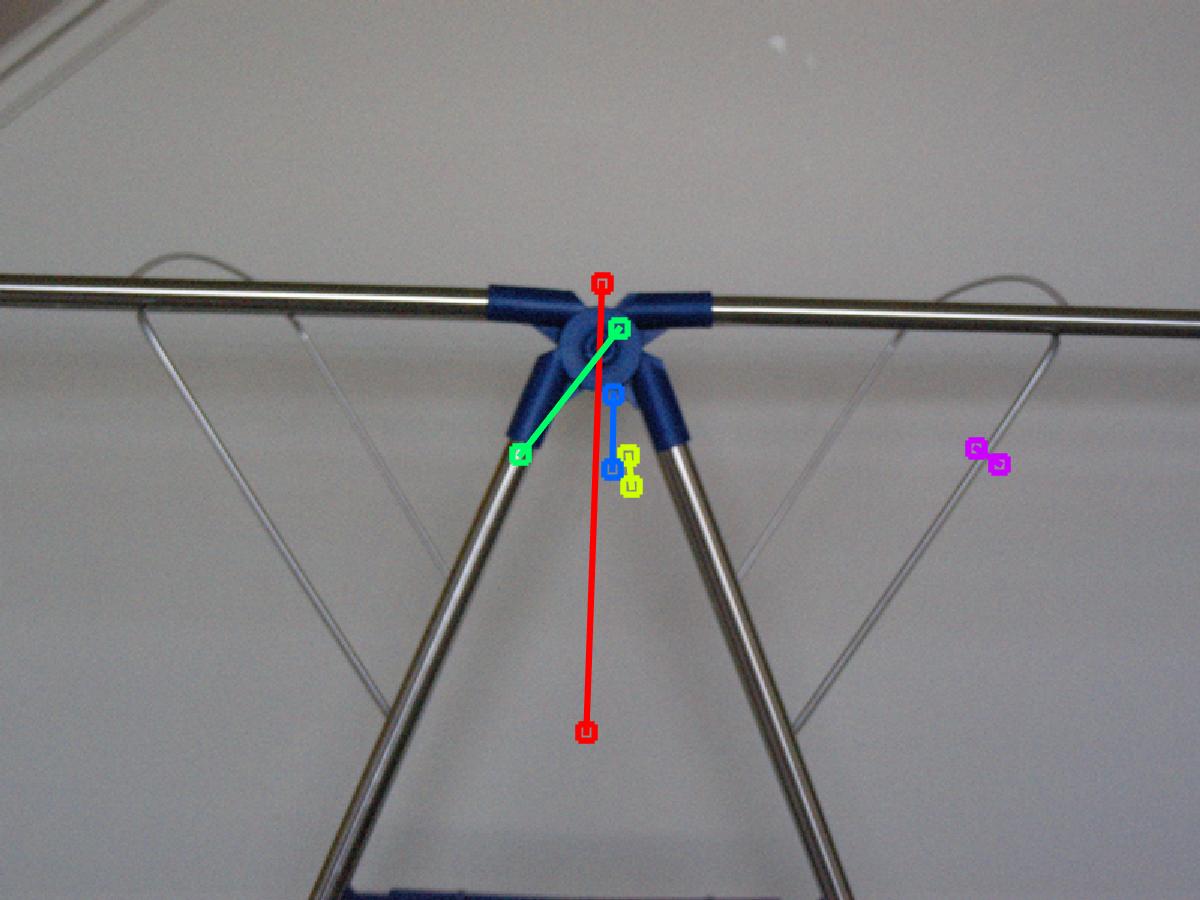}\label{fig:PSU2_Loy} }    
	\subfloat[I012 - Loy2006]{\includegraphics[width=0.33\textwidth]{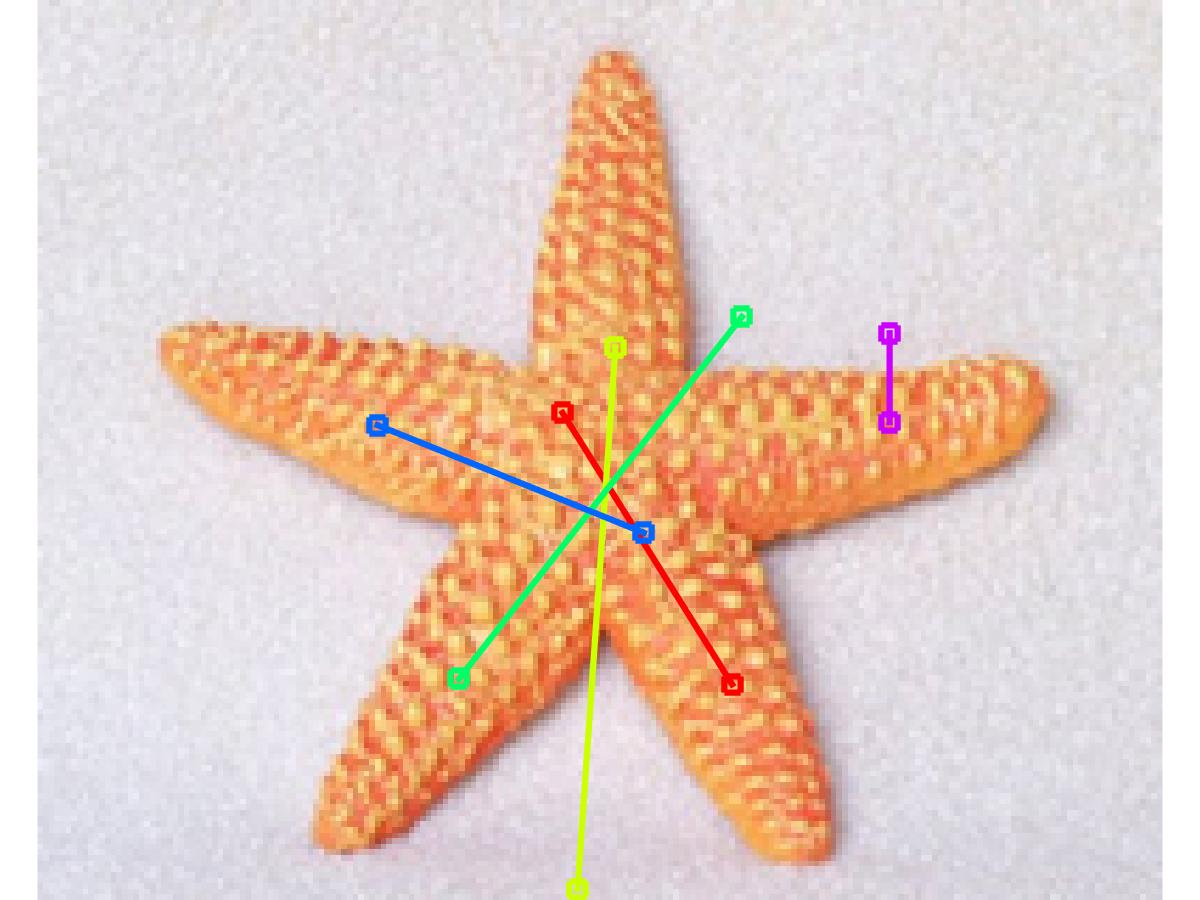}\label{fig:NY1_Loy} } 
	\caption{Some challenging images in PSUm \cite{Rauschert2011,Liu2013} (1st and 2nd columns) and NYm \cite{Cicconet2016} (3rd column) datasets with groundtruth in 1st row (a,b,c). our method in 2nd row (d,e,f) produces better results among El2016 \cite{Elawady2016} in 3rd row (g,h,i) and Loy2006 \cite{Loy2006} in 4th row (j,k,l). For each algorithm, the top five symmetry results is presented in such order: {\color{red} red}, {\color{yellow} yellow}, {\color{green} green}, {\color{blue} blue}, and {\color{magenta} magenta}. Each symmetry axis is shown in a straight line with squared endpoints. Best seen on screen.}
	\label{fig:res}
\end{figure}

  Quantitative (figures~\ref{fig:prCurve},~\ref{fig:MulSymStat}) and qualitative (figure~\ref{fig:res}) comparisons are conducted among the proposed method (Our2017), Loy and Eklundh (Loy 2006) \cite{Loy2006}, Cicconet et al. (Cic2014) \cite{Cicconet2014}, and Elawady et al. (Ela2016) \cite{Elawady2016}. Loy2006 and Ela2016 were reported to have the best performed results for the single symmetry detection in keypoint-based and edge-based methods respectively \cite{Elawady2016}. Figure~\ref{fig:prCurve} presents precision-recall curves for the multiple symmetry datasets (PSUm \cite{Rauschert2011,Liu2013}, and NYm \cite{Cicconet2016}), to compare the proposed method to three prior algorithms (Loy2006 \cite{Loy2006}, Cic2014 \cite{Cicconet2014}, and Ela2016 \cite{Elawady2016}). Cic2014 \cite{Cicconet2014} has the lowest performance for precision and recall in both curves. In figure~\ref{fig:prCurve_PSUm}, Loy2006 \cite{Loy2006} has better precision than Ela2016 \cite{Elawady2016} over corresponding low recall, while the precision of the proposed method (aka Our2017) outperforms all these methods in most sections of the curve. In figure~\ref{fig:prCurve_NYm}, Ela2016 \cite{Elawady2016} has a superior precision performance over Loy2006 \cite{Loy2006} under the recall rate of 40\%, meanwhile the proposed method has the best performance along both precision and recall rates. Additionally, we also compute the $F_1$ score to define the harmonic mean between precision and recall rates, and used the maximum $F_1$ score to qualify the overall performance of different detection algorithms. The values of maximum $F_1$ score are presented in figure~\ref{fig:prCurve} to express the precision-recall curve for each method in a single global measure. Figure~\ref{fig:MulSymStat} shows precision and recall rates where the maximum $F_1$ scores are selected among the corresponding curves in figure~\ref{fig:prCurve}. The proposed method achieved the best performance among all results.


Figure~\ref{fig:res} presents some experimental results for multiple symmetry detection from two publicly available datasets (PSUm \cite{Rauschert2011,Liu2013}, NYm \cite{Cicconet2016}). Groundtruth of the first example (figure~\ref{fig:PSU1_GT}) shows four symmetries (vertical, horizontal, and diagonals) of a flower within natural background view. The proposed method (figure~\ref{fig:PSU1_Our}) detects correctly vertical and horizontal axes, while Loy2006 \cite{Loy2006} (figure~\ref{fig:PSU1_Loy}) fails to find enough feature points resulting two partial (diagonal and vertical) axes, and Ela2016 \cite{Elawady2016} (figure~\ref{fig:PSU1_Ela}) breakdown the vertical axis to represent the top five detections. Second and third examples (figures~\ref{fig:PSU2_GT},~\ref{fig:NY1_GT}) display three in-between symmetries of a thinned metal object and five symmetries expressing arms' details of a starfish respectively over variant texture-less surfaces. These symmetries have been efficiently detected by the proposed method (figures~\ref{fig:PSU2_Our},~\ref{fig:NY1_Our}) over Ela2016 \cite{Elawady2016} (figures~\ref{fig:PSU2_Ela},~\ref{fig:NY1_Ela}). However, Loy2006 \cite{Loy2006} (figures~\ref{fig:PSU2_Loy},~\ref{fig:NY1_Loy}) concentrates on local symmetries describing the inner details of the centric objects.

\section{Conclusion}
This paper proposes a linear-directional kernel-based voting scheme within unified feature representation, in order to support a reliable detection framework for global multiple symmetries. Our approach solves the drawbacks of the previous symmetry detection approaches, by estimating the fixed-sized kernel density with efficient bandwidth parameters, and identifying correctly the symmetrical regions at a global scale. Quantitative and qualitative evaluations present the state-of-the-art performance of our proposed framework among public datasets. This work can be extended to refine the accuracy of the symmetry peaks and the selection of corresponding voting features, using a continuous maxima-seeking technique. The future work is introducing an entropy-based measure, to exploit the global strength of various symmetry axes inside an image.   

\bibliographystyle{splncs03}
\bibliography{ReflectionalSymmetryRefs}

\end{document}